\documentclass{article}

\usepackage{arxiv}

\usepackage[utf8]{inputenc} 
\usepackage[T1]{fontenc}    
\usepackage{hyperref}       
\usepackage{url}            
\usepackage{booktabs}       
\usepackage{amsmath}        
\usepackage{amssymb}        
\usepackage{mathtools}      
\usepackage{bm}             
\usepackage{nicefrac}       
\usepackage{microtype}      
\usepackage{graphicx}       
\usepackage{subcaption}     
\usepackage{float}          
\usepackage{adjustbox}      
\usepackage{tabularx}       
\usepackage{multirow}       
\usepackage{longtable}      
\usepackage{array}          
\usepackage{colortbl}       
\usepackage[dvipsnames]{xcolor}
\definecolor{c1}{HTML}{7bcfa6}
\usepackage{tcolorbox}
\usepackage{lipsum}         
\usepackage{lineno}
\usepackage{amssymb}
\usepackage{amsmath}
\usepackage{array}
\usepackage{doi}            

\hypersetup{
    colorlinks=true, 
    linkcolor=blue,     
    citecolor=blue,   
    urlcolor=blue       
}
\usepackage{fancyhdr}
\newcolumntype{L}[1]{>{\raggedright\arraybackslash}p{#1}}
\newcommand{\figref}[1]{Fig.~\ref{#1}}

\usepackage{caption}
\newtcolorbox{promptbox}[2][Prompt]{
  colback=black!5!white,
  arc=5pt, 
  boxrule=0.5pt,
  fonttitle=\bfseries,
  title=#1, 
  before upper={\small}, 
  fontupper=\fontfamily{ptm}\selectfont,
  colframe=#2, 
}

\title{Lightweight Multimodal Artificial Intelligence Framework for Maritime Multi-Scene Recognition}

\author{ 
    {Xinyu Xi}\thanks{The two authors contribute equally to this work.} \\
    Shanghai Maritime University \\
    Shanghai, China, 201306 \\
    \texttt{xixinyu0160@stu.shmtu.edu.cn} \\
    \And
    {Hua Yang}\thanks{Corresponding author. The two authors contribute equally to this work.} \\
    Shanghai Maritime University \\
    Shanghai, China, 201306 \\
    \texttt{yanghua@shmtu.edu.cn} \\
    \And
    {Shentai Zhang} \\
    Shanghai Maritime University \\
    Shanghai, China, 201306 \\
    \And
    {Yijie Liu} \\
    Shanghai Maritime University \\
    Shanghai, China, 201306 \\
    \And
    {Sijin Sun} \\
    National University of Singapore \\
    Singapore, 117597 \\
    \texttt{sun.sijin@u.nus.edu} \\
    \And
    {Xiuju Fu}\thanks{Corresponding author.} \\
    Institute of High Performance Computing, A*STAR \\
    Singapore, 138634 \\
    \texttt{fuxj@ihpc.a-star.edu.sg} \\
}

\renewcommand{\shorttitle}{}

\hypersetup{
pdftitle={A template for the arxiv style},
pdfsubject={q-bio.NC, q-bio.QM},
pdfauthor={David S.~Hippocampus, Elias D.~Striatum},
pdfkeywords={First keyword, Second keyword, More},
}

\begin{document}
\maketitle

\begin{abstract}
Maritime Multi-Scene Recognition is crucial for enhancing the capabilities of intelligent marine robotics, particularly in applications such as marine conservation, environmental monitoring, and disaster response. However, this task presents significant challenges due to environmental interference, where marine conditions degrade image quality, and the complexity of maritime scenes, which requires deeper reasoning for accurate recognition. Pure vision models alone are insufficient to address these issues. To overcome these limitations, we propose a novel multimodal Artificial Intelligence (AI) framework that integrates image data, textual descriptions and classification vectors generated by a Multimodal Large Language Model (MLLM), to provide richer semantic understanding and improve recognition accuracy. Our framework employs an efficient multimodal fusion mechanism to further enhance model robustness and adaptability in complex maritime environments. Experimental results show that our model achieves 98$\%$ accuracy, surpassing previous SOTA models by 3.5$\%$. To optimize deployment on resource-constrained platforms, we adopt activation-aware weight quantization (AWQ) as a lightweight technique, reducing the model size to 68.75MB with only a 0.5$\%$ accuracy drop while significantly lowering computational overhead. This work provides a high-performance solution for real-time maritime scene recognition, enabling Autonomous Surface Vehicles (ASVs) to support environmental monitoring and disaster response in resource-limited settings.
\end{abstract}


\section{Introduction}
In recent years, the rapid advancement of intelligent marine robotics has significantly enhanced the capabilities of Autonomous Surface Vehicles (ASVs), particularly in autonomous navigation, intelligent perception, and remote control. These advancements have broadened the scope of ASVs in critical maritime applications such as marine conservation, environmental monitoring and disaster response. By leveraging cutting-edge technologies, ASVs are increasingly integral to ensuring efficient and reliable operations in resource-limited environments, contributing to both the sustainability of marine ecosystems and the safety of maritime operations~\cite{yuan2023marine,wu2023cooperative,bella2021hmdcs,sun2025self}. However, the increasing complexity of marine environments necessitates robust and accurate recognition algorithms to address the challenges of diverse marine scenes.

Efficiently addressing maritime tasks requires significant advancements in Artificial Intelligence (AI), particularly deep learning (DL) techniques~\cite{10.3389/fevo.2023.1257542, 10.2112/JCR-SI112-113.1}. While deep learning architectures such as Convolutional Neural Networks (ConvNeXt), Residual Networks (ResNets), EfficientNets, and Vision Transformers (ViTs)~\cite{liu_convnet_2022,he_deep_2016,tan_efficientnet_2019,dosovitskiy_image_2020} have led to substantial improvements in terrestrial applications like urban surveillance and habitat monitoring, the marine environment introduces unique challenges that require further innovation. Factors such as dynamic lighting, water surface reflections, occlusions, and the increasing intensity of human activities in coastal regions highlight the pressing need for effective scene recognition solutions. These solutions are essential for ensuring the efficiency and safety of ASVs in supporting environmental protection and maritime operations.

To tackle the challenges posed by complex maritime environments, we propose a novel multimodal AI framework that combines image features, textual descriptions, and classification vectors. These textual descriptions and classification vectors are generated by a Multimodal Large Language Model (MLLM), which leverages its advanced reasoning capabilities to provide rich semantic context based on input images, significantly enhancing the feature set and improving recognition accuracy.

One of the core aspects of our framework is an efficient feature extraction process that employs state-of-the-art technologies tailored for each modality. The Swin Transformer extracts fine-grained image features, BERT processes textual data, and a Multi-Layer Perceptron (MLP) handles classification vectors. These techniques ensure that each modality contributes its most relevant information, strengthening the overall feature set and ensuring robustness.

Our framework employs a sophisticated multimodal fusion mechanism with four key components: attention mechanisms, weighted integration, enhanced modal alignment, and dynamic modality prioritization. These strategies enhance feature relevance, balance modality contributions, align modalities in a shared space, and dynamically adjust priority based on the scene, improving robustness and making it ideal for real-time deployment in resource-constrained maritime environments like ASVs.

To further enhance the framework’s efficiency, we incorporate Activation-aware Weight Quantization (AWQ)~\cite{lin2024awq} as a lightweight technique, optimizing the model for deployment in resource-constrained environments. This strategy reduces the model size and computational overhead while maintaining high performance, making it ideal for real-time maritime applications.

In summary, our framework combines advanced deep learning models with multimodal fusion techniques to address the unique challenges of maritime scene recognition. Its online training and offline deployment ensures operational effectiveness in remote environments. This approach sets the foundation for the future development of intelligent marine robotics, enabling more reliable and efficient systems for environmental monitoring, disaster response, and other maritime applications.

\section{Related Work}
\subsection{Multiscene Recognition}

Multiscene recognition is a critical task in computer vision, aimed at analyzing image content and assigning semantic labels to help computers understand different environments and contexts. Unlike traditional object recognition tasks, multiscene recognition not only focuses on objects within an image but also analyzes the semantic relationships between objects and background information to identify the type of scene represented, such as indoor, urban, marine, or forest scenes. Multiscene recognition faces challenges such as the complexity of scene content, the similarity between different scenes, and semantic ambiguity, which make it difficult to differentiate visually similar scenes with varying semantic meanings. Additionally, improving model robustness and adaptability while maintaining high recognition accuracy is a key challenge~\cite{XIE2020}.

For example, the Local Semantic Enhanced Convolutional Network (LSE-Net) successfully improved the recognition accuracy of aerial scenes by simulating local region relationships in human vision~\cite{BI2021}. In the field of remote sensing scene recognition, the Resource-Efficient Attention Network (RTANet) enhanced model performance in diverse scenes by capturing long-range dependencies in remote sensing images~\cite{FU2020}. The Spatial-Channel Transformer (SC-Transformer) significantly improved fine-grained classification capabilities by integrating spatial and channel information~\cite{BAIK2022}. Additionally, the Adaptive Local Recalibration Network (ALR-Net) improved model adaptability to multiscene recognition through adaptive data augmentation, especially excelling in capturing local information~\cite{WANG2023}.

Although many methods have achieved success in multiscene recognition, our research focuses on marine scene recognition, specifically for applications in intelligent marine robotics. Building upon deep learning, our approach integrates multimodal fusion techniques to enhance the model’s robustness and recognition accuracy across different marine environments. Through these innovations, our approach not only demonstrates outstanding performance in multiscene recognition tasks but also achieves efficient and precise recognition in specific marine environment monitoring scenarios.

\subsection{Multimodal Image Recognition}
\subsubsection{Traditional Multimodal Image Recognition}

Traditional unimodal image recognition methods, which rely solely on visual data, such as CLIP~\cite{radford2021learning} and BLIP~\cite{li2022blip}, often face significant limitations in understanding complex scenes. Images alone may lack sufficient semantic context, and crucial information from other modalities, such as text or signals, is typically underutilized. To address these gaps, multimodal image recognition integrates multiple data sources, such as images, textual descriptions, and signals, e.g., audio, time-series data, enhancing the model's ability to understand and interpret more complex real-world scenarios. This integration improves accuracy and robustness by providing richer contextual information that would be missed in unimodal approaches.

In traditional multimodal systems, different types of data are processed using independent encoders. For example, textual data is typically processed using pre-trained language models, e.g., BERT to extract semantic features, while images are encoded using CNNs or ViTs to capture visual features~\cite{li2021towards, likhosherstov2021polyvit}. Signals or time-series data, such as audio or sensor inputs, are often encoded using recurrent neural networks (RNNs) or temporal convolutions, depending on their nature, discrete or continuous~\cite{zhou2024dpnet}. Once each modality is processed, the features are fused, often through methods like early fusion, late fusion, or attention-based fusion, to combine the complementary information from different sources~\cite{lu2019vilbert, tan2019lxmert, labbakiorthogonal}. This fusion allows the model to generate a more holistic understanding of the scene.

Despite these advances, traditional multimodal methods still struggle with semantic alignment and cross-modal understanding. While they improve recognition by integrating various sources of information, they often lack the deep semantic capabilities and reasoning provided by Large Language Models (LLMs). This limitation hinders their ability to fully leverage contextual cues, particularly in complex or ambiguous tasks.

\subsubsection{Multimodal Image Recognition Technology With LLMs}
To address this, modern multimodal recognition systems integrate vision encoders with LLMs, enabling a seamless fusion of visual and textual features. By leveraging the reasoning power of LLMs, these systems enhance cross-modal alignment and provide a deeper, more context-aware fusion, significantly improving recognition performance in challenging scenarios. For instance, methods like LLaVA~\cite{liu2023visual} and MiniGPT-4~\cite{zhu2023minigpt} align the visual information extracted by the vision encoder with the reasoning and semantic capabilities of the LLM. In these approaches, the LLM plays a crucial role in interpreting the visual data in a more structured and contextually relevant way, allowing the model to make informed decisions based on both visual and linguistic cues. This integration results in improved performance, particularly in tasks that involve complex scenarios or require reasoning, such as fine-grained object recognition, scene understanding, or ambiguity resolution. By incorporating the reasoning power of LLMs, these systems move beyond traditional image recognition capabilities and become more adaptable and accurate in real-world applications. However, multimodal image recognition integrating large language models still faces challenges. The main issues include high computational complexity, especially for real-time processing, and difficulty in balancing reasoning capabilities with efficiency. Additionally, effectively integrating multimodal data and minimizing noise remain challenging, particularly in tasks involving complex cross-modal interactions, open-vocabulary recognition, and long-tail problems. These are precisely the challenges that Our framework aim to address.

\section{Methodology}
\subsection{Overview}
We propose a novel multimodal scene recognition method for the task of maritime scene recognition, which integrates multiple modalities to improve recognition accuracy and robustness. This section introduces the two key components of our proposed method: data processing, model design and lightweight design. Data processing ensures input consistency for multimodal fusion, serving as a crucial prerequisite for efficient model operation. The model design is centered on efficient feature extraction and fusion, aiming to address the challenges posed by diverse maritime environments and constrained computational resources. Additionally, the lightweight design through AWQ enhances the model's deployment efficiency, significantly reducing the model size and computational complexity without compromising accuracy.

\subsection{Data Description and Processing}

The dataset utilized in this study was meticulously curated to replicate real-world conditions encountered by ASVs in marine scene recognition tasks. It comprises five distinct categories: marine debris, animal stranding, ship fire, ship capsize, and red tide. Each category contains 100 images, resulting in a total of 500 images. This diverse dataset serves as a robust foundation for training and evaluating multimodal recognition systems in maritime environments, integrating three complementary modalities: Image Data, Textual Data, and Vector Data. The latter two modalities are generated by a MLLM based on the corresponding images, enabling effective multimodal learning.

\subsubsection{Image Data}
\label{subsubsec:3.2.1}
Images were acquired from global news platforms, environmental agencies, and open-access marine repositories, ensuring a comprehensive representation of real-world scenarios. The dataset contains images under various challenging environmental conditions, such as dynamic water surfaces and changing lighting, with images taken from low-altitude and near-water perspectives, typically captured by ASVs.

For processing, to ensure consistency and minimize environmental variations, all images are resized to a uniform $224 \times 224$ resolution. Additionally, the pixel values are normalized by adjusting each color channel (RGB) using predefined mean and standard deviation values. This normalization process stabilizes the data, improving model training efficiency and ensuring better convergence. The processed images are denoted as $\mathbf{I}$, optimizing them for robust learning and accurate model performance.

\subsubsection{Textual Data} 
\label{subsubsec:3.2.2}
To enable multimodal learning, each image in the dataset is paired with a corresponding textual description generated by a MLLM. These descriptions provide semantic context for the visual data, enriching the dataset’s information content and supporting feature extraction across modalities. ~\figref{fig:dataset_examples} shows the representative images of each category and their corresponding textual descriptions.

Textual descriptions are first tokenized and then either truncated or padded to a predefined maximum length, resulting in a sequence denoted as $T$, which is subsequently processed through a transformer to obtain a representation.

\subsubsection{Vector Data} 
\label{subsubsec:3.2.3}
Each image is paired with a probabilistic classification vector generated by the MLLM, representing the likelihood of the image belonging to one of five categories: \texttt{Red Tide}, \texttt{Marine Debris}, \texttt{Animal Stranding}, \texttt{Ship Fire}, or \texttt{Ship Capsize}. These vectors provide valuable probabilistic information, supporting multimodal learning and enhancing feature extraction. 

The classification vectors are first retrieved from a list, converts it from a string representation back into its original format, and then transforms it into a numerical tensor \( V_{0} \). This ensures that the vector data is in the correct format, ready for further processing and integration into the model.

\begin{figure}[H]
\centering
\includegraphics[width=\linewidth]{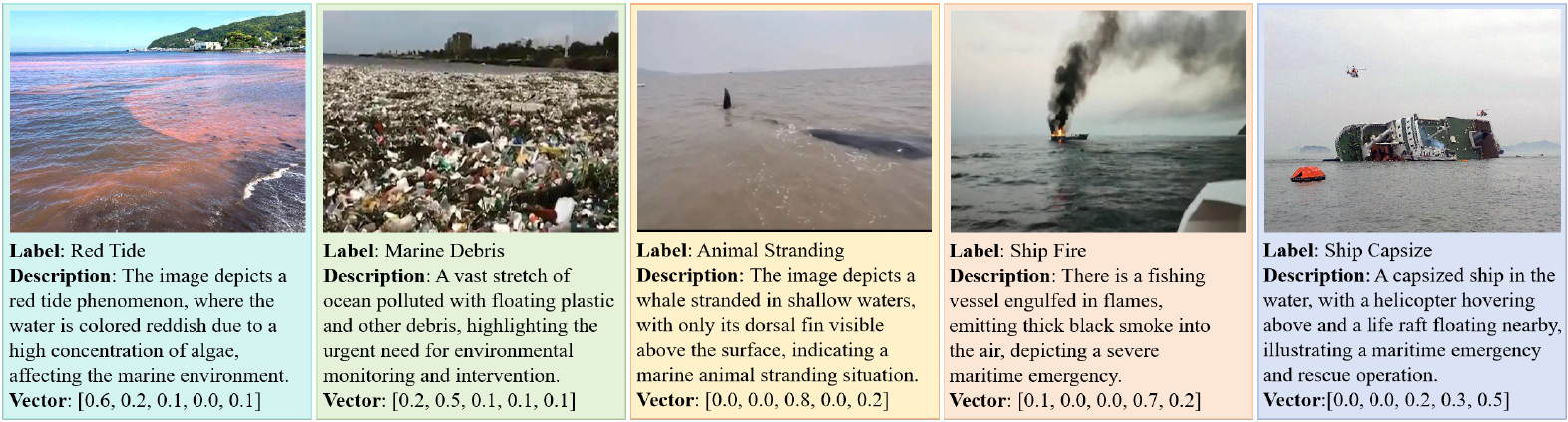}
\caption{Examples of marine scene categories in the dataset: marine debris, animal stranding, ship fire, ship capsize, and red tide. All images simulate low-altitude, near-water perspectives, providing realistic scenarios for training multimodal recognition systems.}\label{fig:dataset_examples}
\end{figure}

In conclusion, ~\figref{fig:dataset_examples} showcases a subset of the dataset. By combining images, textual descriptions, and classification vectors, the dataset captures complementary features for robust classification. This well-curated dataset forms a solid foundation for advancing ocean scene recognition and real-world applications.

The processing strategy optimizes the dataset by resizing and normalizing images, tokenizing and embedding textual descriptions, and normalizing classification vectors for consistency. This streamlined pipeline enhances multimodal feature extraction and fusion, improving model generalization and supporting effective integration for maritime scene recognition.

\subsection{Framework}
\begin{figure}[H]
\centering
\includegraphics[width=\linewidth]{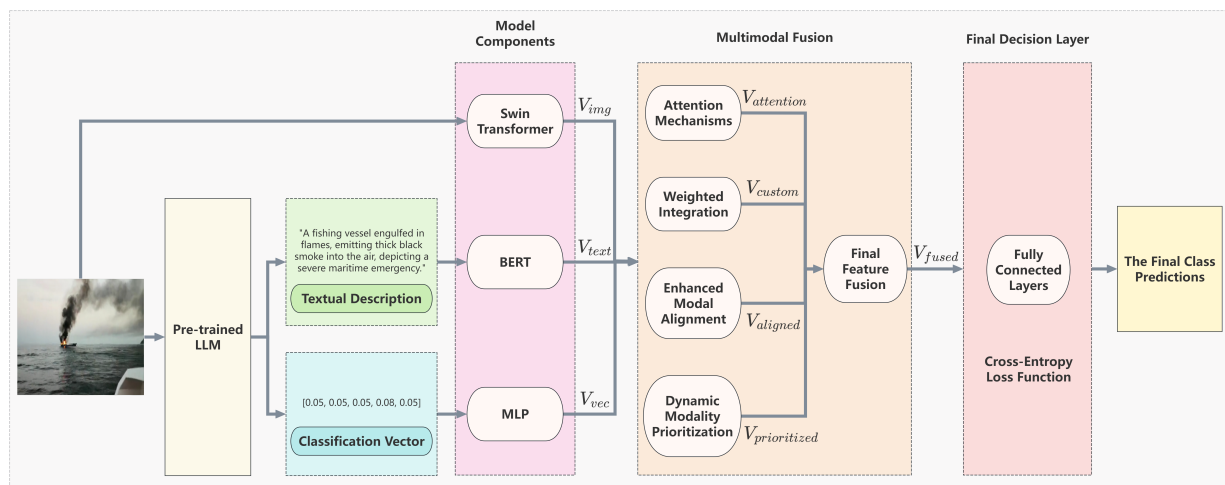}
\caption{Framework diagram of the multimodal marine scene recognition system.}\label{frame}
\end{figure}

The specific design framework of the model is shown in ~\figref{frame}.

\subsubsection{Model Component In Feature Extraction}

\paragraph{\textbf{A}}\textbf{Swin Transformer for Image Processing:}

For the input image \( I \) in~\ref{subsubsec:3.2.1} with dimensions \( B \times C \times H \times W \), where \( B \) is the batch size, \( C \) is the number of channels, \( H \) is the height, and \( W \) is the width, we first divide the image into non-overlapping patches of size \( P \times P \). The number of patches, \( N \), is determined by the image dimensions \( H \) and \( W \) divided by \( P \). Each patch is then embedded into a high-dimensional space, resulting in a sequence of patch embeddings \( X_{0} \in \mathbb{R} ^{B \times N \times D_X }\), where \( D_X \) is the embedding dimension. The sequence of embedded patches is processed through Swin Transformer blocks with shifted window-based multi-head self-attention (SW-MSA):

\begin{equation}
X_{out} = \text{SwinTransformer}(X_{0})
\end{equation}
where \( X_{out} \) denotes the final layer output.

\paragraph{\textbf{B}}\textbf{BERT for Textual Data Processing:}

The textual data is processed using a pre-trained BERT model, which extracts contextual semantic features from the processed text input \( T \) in~\ref{subsubsec:3.2.2}. \( T \) is first tokenized and mapped to a continuous embedding space:

\begin{equation}
T_0 = \text{Embed}(T) 
\end{equation}
where \( T_0 \) represents the tokenized input with dimensions \( B \times L \times D_T \), where \( L \) is the sequence length and \( D_T \) is the embedding dimension.

The embedded sequence \( T_0 \) is passed through multiple self-attention layers of the BERT model, yielding a contextualized representation \( T_{out} \):

\begin{equation}
T_{out} = \text{BERT}(T_0) 
\end{equation}
where \( T_{out} \) is the output of the final layer of the BERT model. 

\paragraph{\textbf{C}}\textbf{MLP for Vector Data Processing:}

For processing the classification vectors \( V_{0} \) in~\ref{subsubsec:3.2.3}, we utilize a two-layer Multi-Layer Perceptron (MLP). The input classification vector \( V_{0} \) is passed through the MLP to project it into a feature space, as described by:

\begin{equation}
V_{\text{out}} = \phi_V(W_2 \sigma(W_1 V_{0} + b_1) + b_2) 
\end{equation}
where \( W_1 \), \( W_2 \), \( b_1 \), and \( b_2 \) are learnable parameters, and \( \sigma \) is the ReLU activation function. The projection function \( \phi_V \) maps the resulting vector into the desired feature space. The output, \( V_{\text{out}} \), represents the processed classification vector, which is then used in the multimodal fusion process.

\subsubsection{Multimodal Fusion}
To ensure that the features from the three modalities—image, text, and classification vectors—are aligned in the same feature space for subsequent fusion, we perform linear transformations to map them to a common dimensionality. Specifically, the image features \( V_{\text{img}} \in \mathbb{R}^{B \times d} \), text features \( V_{\text{text}} \in \mathbb{R}^{B \times d} \), and classification vector features \( V_{\text{vec}} \in \mathbb{R}^{B \times d} \) are processed through the following transformations:

\begin{equation}
\begin{split}
V_{\text{img}} &= W_{\text{img}} X_{\text{out}} + b_{\text{img}}, \\
V_{\text{text}} &= W_{\text{text}} T_{\text{out}} + b_{\text{text}}, \\
V_{\text{vec}} &= V_{\text{out}}
\end{split}
\end{equation}
where \( W_{\text{img}} \), \( W_{\text{text}} \) are the weight matrices for the image and text features, and \( b_{\text{img}} \), \( b_{\text{text}} \) are the corresponding bias terms. After these transformations, the features from all three modalities are aligned in the same dimensional space, ready for further fusion.

\paragraph{\textbf{A}}\textbf{Attention Mechanisms:}

For attention mechanisms, \( V_{\text{stack}} \) is formed by concatenating \( V_{\text{img}} \), \( V_{\text{text}} \), and \( V_{\text{vec}} \) along the feature dimension to create a stacked feature matrix. This stacked matrix \( V_{\text{stack}} \) is then passed through a self-attention mechanism. First, we compute the query (\( Q \)), key (\( K \)), and value (\( V' \)) matrices using the learned weight matrices \( W_Q \), \( W_K \), and \( W_V \), respectively:

\begin{equation}
Q = V_{\text{stack}} W_Q, \quad K = V_{\text{stack}} W_K, \quad V' = V_{\text{stack}} W_V
\end{equation}

The similarity between the query and key is then computed to obtain the attention weight matrix \( A \):

\begin{equation}
A = \text{softmax}\left(\frac{Q K^T}{\sqrt{d_k}}\right)
\end{equation}
where \( d_k \) is the dimension of the keys.

Finally, the attention weight matrix \( A \) is used to calculate the output feature matrix \( V_{\text{attention}} \) by multiplying \( A \) with the value matrix \( V' \):

\begin{equation}
V_{\text{attention}} = A V'
\end{equation}

\paragraph{\textbf{B}}\textbf{Weighted Integration:}

After applying the attention mechanism, we move to weighted integration to further refine the fusion of features. This step optimizes the combination of image, text, and classification vectors, adjusting the weight of each modality based on available resources and their relevance to the task: \begin{linenomath} \begin{equation} V_{\text{custom}} = \alpha V_{\text{img}} + \beta V_{\text{text}} + \gamma V_{\text{vec}}, \end{equation} \end{linenomath} where $\alpha$, $\beta$, and $\gamma$ are learnable coefficients, which dynamically adjust based on the available resources and their contribution to the classification task.

\paragraph{\textbf{C}}\textbf{Enhanced Modal Alignment for Edge Inference:}

To enhance the alignment between the features from different modalities, we first utilize two methods: maximizing mutual information (MI) and minimizing Jensen-Shannon (JS) divergence. These techniques help to align image features, textual features, and classification vectors in a shared representation space, facilitating more effective multimodal fusion.

Mutual information (MI) quantifies the shared information between two random variables. To align image features \( V_{\text{img}} \) and text features \( V_{\text{text}} \), we aim to maximize their mutual information. This process enhances the relevance between the two modalities and strengthens their consistency in the shared feature space.
          
Given the image features \( V_{\text{img}} \) and text features \( V_{\text{text}} \), the mutual information between them is defined as:

\begin{equation}
I(V_{\text{img}}; V_{\text{text}}) = H(V_{\text{img}}) - H(V_{\text{img}} | V_{\text{text}})
\end{equation}
where \( H(V_{\text{img}}) \) is the entropy of the image features, representing the distribution of the image data, and \( H(V_{\text{img}} | V_{\text{text}}) \) is the conditional entropy of the image features given the text features, representing the distribution of image features when text is available.

The goal is to maximize the mutual information, \( I(V_{\text{img}}; V_{\text{text}}) \), which leads to the following objective function for MI maximization:

\begin{equation}
L_{\text{MI}} = -I(V_{\text{img}}; V_{\text{text}}) = H(V_{\text{img}}) - H(V_{\text{img}} | V_{\text{text}})
\end{equation}

By maximizing mutual information, we enhance the correlation between image and text features, improving their alignment. The alignment function is defined as:

\begin{equation}
V_{\text{aligned}}^{\text{MI}} = f_{\text{align}}^{\text{MI}}(V_{\text{img}}, V_{\text{text}}) = \text{argmax}_{V_{\text{img}}, V_{\text{text}}} \mathcal{L}_{\text{MI}}
\end{equation}

Jensen-Shannon (JS) divergence measures the similarity between two probability distributions. To make the distributions of image and text features more consistent, we minimize their JS divergence. The JS divergence is formulated as:

\begin{equation}
L_{\text{JS}} = \text{JS}(P(V_{\text{img}}, V_{\text{text}}) \parallel P(V_{\text{img}})P(V_{\text{text}}))
\end{equation}
where \( P(V_{\text{img}}, V_{\text{text}}) \) is the joint distribution of image and text features, and \( P(V_{\text{img}}) \) and \( P(V_{\text{text}}) \) are the marginal distributions of image and text features, respectively.

We aim to minimize \( L_{\text{JS}} \) to make the joint distribution of image and text features as close as possible to their marginal distributions, reducing redundancy between modalities. The alignment expression for JS divergence minimization is:

\begin{equation}
V_{\text{aligned}}^{\text{JS}} = f_{\text{align}}^{\text{JS}}(V_{\text{img}}, V_{\text{text}})
\end{equation}

To leverage the strengths of both MI maximization and JS divergence minimization, we combine these two objectives into a compound loss function. This allows us to simultaneously maximize the relevance between image and text features while minimizing their redundancy.

The combined loss function is given by:

\begin{equation}
L_{\text{align}} = \lambda_{\text{MI}} \cdot L_{\text{MI}} + \lambda_{\text{JS}} \cdot L_{\text{JS}}
\end{equation}
where \( \lambda_{\text{MI}} \) and \( \lambda_{\text{JS}} \) are hyperparameters that control the relative importance of each objective.

Finally, by minimizing this compound loss, we obtain the aligned features \( V_{\text{aligned}} \), as follows:

\begin{equation}
V_{\text{aligned}} = f_{\text{align}}(V_{\text{img}}, V_{\text{text}}, V_{\text{vec}})= \lambda_{\text{MI}} \cdot f_{\text{align}}^{\text{MI}}(V_{\text{img}}, V_{\text{text}}) + \lambda_{\text{JS}} \cdot f_{\text{align}}^{\text{JS}}(V_{\text{img}}, V_{\text{text}})
\end{equation}

\paragraph{\textbf{D}}\textbf{Dynamic Modality Prioritization:}

To dynamically adjust the contribution of each modality in the fusion process, we calculate a priority score \( P_m \) for each modality. This score determines the relative weight of each modality, allowing the model to focus on the most informative modality at different stages of processing.

We first compute a priority score \( P_m \) for each modality \( m \) (where \( m \in \{\text{img}, \text{text}, \text{vec}\} \)) to quantify its importance in the fusion process. The formula for calculating the priority score is as follows:

\begin{equation}
P_m = w_m \cdot \text{Relevance}(V_m)
\end{equation}
where \( P_m \) is the priority score for modality \( m \), \( V_m \) is the original feature representation for modality \( m \), \( w_m \) is a learnable weight for each modality, representing its global importance, \( \text{Relevance}(V_m) \) denotes the relevance of modality \( m \) in the current task, which is dynamically computed through the model's training process based on classification accuracy.

Once the priority score for each modality is calculated, the contribution of each modality is dynamically adjusted based on its priority score. Specifically, the output features of each modality are multiplied by its respective priority score, ensuring that higher-priority modalities have a larger influence on the final output. This adjustment is represented as:

\begin{equation}
V_{\text{prioritized}} = \sum_{m \in \{\text{img}, \text{text}, \text{vec}\}} P_m \cdot V_m
\end{equation}

By multiplying each modality’s original features by its priority score \( P_m \), and then summing the adjusted features across all modalities, we obtain the fused feature representation \( V_{\text{prioritized}} \). This process ensures that modalities with higher relevance contribute more significantly to the final multimodal feature, allowing the model to dynamically focus on the most informative input data.

\paragraph{\textbf{E}}\textbf{Final Feature Fusion:}

At this point, the outputs of Enhanced Modal Alignment and Dynamic Modality Prioritization are incorporated into the final fused feature vector: \begin{linenomath} \begin{equation} V_{\text{fused}} = \phi_F\left(V_{\text{attention}} \oplus V_{\text{custom}} \oplus V_{\text{aligned}} \oplus V_{\text{prioritized}}\right) \end{equation} \end{linenomath} where $V_{\text{aligned}}$ refers to the aligned features from the Enhanced Modal Alignment step, and $V_{\text{prioritized}}$ refers to the prioritized features from Dynamic Modality Prioritization. The $\oplus$ operator denotes concatenation, and $\phi_F$ represents the final transformation layer that aligns the fused features with the classification task.

\subsubsection{Final Decision Layer}

The final step of the model maps the fused features to marine scene categories using fully connected layers, followed by a ReLU activation function. The model is optimized with a cross-entropy loss function.

\paragraph{\textbf{A}}\textbf{Fully Connected Layers:}

The fully connected layers integrate the fused features into a decision space. The first layer reduces the feature dimensions, and the second layer projects the reduced representation into the final output space, where the dimensionality corresponds to the number of classes. The transformations can be expressed as:

\begin{equation}
\mathbf{H_2} = W_2 \sigma(W_1 V_{\text{fused}} + b_1) + b_2
\end{equation}
where \( \sigma \) represents the ReLU activation function.

The output \( \mathbf{H_2} \) contains the probabilities for each of the \( n \) classes, which is passed through a softmax layer to generate the final class predictions.

\paragraph{\textbf{B}}\textbf{Cross-Entropy Loss Function:}

The final output is passed through the second fully connected layer, resulting in a probability distribution over the $n$ classes. The classification loss is computed using the cross-entropy loss function, which is a standard loss function for classification tasks. It calculates the difference between the true labels and the predicted probabilities, providing a measure of how well the model is performing. The formula for the cross-entropy loss is:
\begin{linenomath}
\begin{equation}
\mathcal{L}_{\text{class}} = -\sum_{i=1}^{n_{\text{class}}} y_i \log \hat{y}_i,
\end{equation}
\end{linenomath}
where $y_i$ represents the true label (one-hot encoded) for class $i$, \( \hat{y}_i \) is the predicted probability for class \( i \) after applying the Softmax function, where \( \hat{y}_i = \frac{e^{H_{2i}}}{\sum_{j=1}^{n_{\text{class}}} e^{H_{2j}}} \), and \( H_{2i} \) is the output of the second fully connected layer for class \( i \).

This comprehensive approach ensures that the model can effectively map the fused multimodal features to the appropriate marine scene categories while optimizing for accuracy through backpropagation.

\subsection{AWQ for Lightweight Model Deployment}

\begin{figure}[H]
\centering
\includegraphics[width=\linewidth]{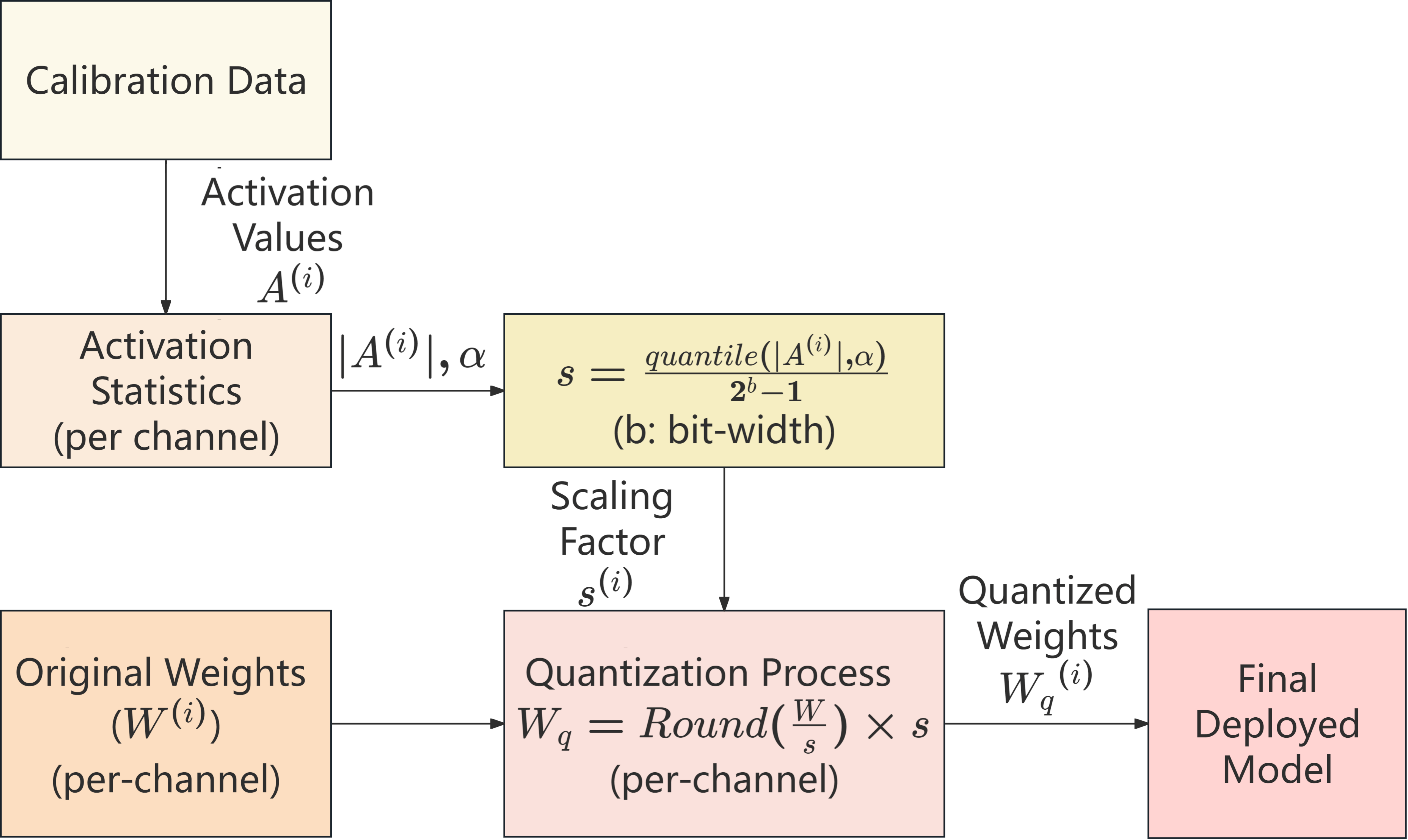}
\caption{The post-training quantization process using AWQ for efficient deployment.}\label{AWQ}
\end{figure}

To enhance the efficiency of our multimodal maritime scene recognition model, we apply AWQ~\cite{lin2024awq} as a post-training quantization (PTQ) method. AWQ applies PTQ by dynamically adjusting the quantization scales based on activation distributions, ensuring minimal accuracy loss while significantly decreasing model size. Unlike traditional uniform quantization, AWQ assigns higher precision to weight channels corresponding to larger activation magnitudes, ensuring critical features retain high fidelity. This quantization strategy shown in ~\figref{AWQ} is applied to various components of the model, including the Swin Transformer (for image processing), BERT (for text processing), MLP (for classification vector processing), and the attention-based fusion module. Below, we detail how AWQ is applied to each of these components and describe the calibration process used for efficient deployment.

\subsubsection{Quantization of Model Components}

\paragraph{\textbf{A}}\textbf{Swin Transformer (Image Processing):}

For the Swin Transformer layers, particularly the Feed-Forward Networks (FFN) and attention projections, per-channel weight quantization is applied to maintain computational efficiency while preserving critical feature representations. The scaling factor for each channel is computed from activation statistics obtained from a calibration dataset, ensuring that channels with higher activation magnitudes retain greater precision. This prevents excessive quantization errors in key attention operations.

To implement this, activation statistics are collected from the outputs of the SW-MSA and FFN layers. The per-channel scaling factor is derived using a high quantile of the absolute activation values, allowing the model to allocate more bit precision to channels that have a greater impact on feature representations. This strategy helps Swin Transformer maintain strong generalization in maritime scene recognition tasks, despite the lower-bit representation.

\paragraph{\textbf{B}}\textbf{BERT (Text Processing):}

For the self-attention layers in BERT, AWQ is applied to the key and value projections, ensuring that important semantic relationships between words are preserved while reducing memory consumption. Unlike traditional post-training quantization methods, AWQ dynamically adjusts the per-channel scaling factors based on activation distributions collected after layer normalization (LN).

Since LayerNorm significantly alters activation distributions, computing scaling factors before normalization could lead to incorrect quantization ranges, causing instability during inference. To mitigate this, we extract activation statistics from post-LN activations, aligning the quantization parameters with the actual inference-time activation ranges.

Additionally, the LayerNorm and Softmax operations are kept in full precision to prevent numerical instability, particularly in low-bit representations where floating-point precision errors can accumulate in gradient updates.

\paragraph{\textbf{C}}\textbf{MLP (Classification Vector Processing):}

For the MLP layers that process classification vectors, per-channel AWQ quantization is applied to preserve the integrity of the classification outputs while reducing storage and computational costs. Since these classification vectors contain probability-based representations that influence decision-making, their distributions differ significantly from raw image or text embeddings.

To ensure effective quantization, the activation statistics are collected from the outputs of the classification vector layer, rather than the input embeddings. The scaling factor for each channel is computed using the high quantile of activation values, ensuring that important probability scores or confidence values are not overly compressed. This method reduces the risk of quantization-induced misclassification while maintaining the efficiency of low-bit inference.

\paragraph{\textbf{D}}\textbf{Attention-Based Fusion Module:}

In the fusion module, which integrates features across modalities, the weight matrices for queries (\( Q \)), keys (\( K \)), and values (\( V \)) projections are quantized per-channel. The scaling factors for these projections are computed independently based on activation distributions to preserve the dynamic range of the attention mechanism.

A critical aspect of quantization in this module is maintaining the balance between modalities during feature aggregation. Since different modalities have inherently different activation distributions, e.g., image embeddings may have higher variance than textual embeddings, separate scaling factors are computed for each modality before fusion.

To avoid instability in the attention weights, the Softmax operation remains in full precision. This prevents small numerical errors in low-bit precision computations from disproportionately affecting attention weight calculations, ensuring a stable multimodal representation.

\subsubsection{Unified AWQ Quantization Formulation}

For each model component, weights are quantized per-channel using activation-aware scaling:

\begin{linenomath}
\begin{equation}
\begin{aligned}
W_q^{(i)} &= \text{Round}\left( \frac{W^{(i)}}{s^{(i)}} \right) \times s^{(i)}, \\
s^{(i)} &= \frac{\text{quantile}(|A^{(i)}|, \alpha)}{2^{b}-1}
\end{aligned}
\end{equation}
\end{linenomath}
where \( W^{(i)} \) and \( W_q^{(i)} \) denote the original and quantized weights for the \( i \)-th channel. \( A^{(i)} \) represents activation values collected during calibration. \( s^{(i)} \) is the per-channel scaling factor, computed using the \( \alpha \)-th quantile of activation values. \( b \) is the quantization bit-width. Softmax and LayerNorm operations remain in full precision to ensure stability.

\subsubsection{Calibration and Deployment}

AWQ is a PTQ method, meaning that the model is quantized after training without requiring additional fine-tuning or gradient updates. This makes AWQ highly efficient for deploying large-scale models in resource-constrained environments, such as ASVs, where computational power and memory are limited.

During the calibration phase, activation statistics are collected from randomly sampled training instances, capturing the real-world distribution of activations across different layers. These statistics are then used to compute per-channel scaling factors, ensuring that channels with higher activation magnitudes are assigned finer quantization precision, thereby minimizing information loss. This approach differs from traditional uniform quantization by dynamically adjusting the quantization range per channel, preserving critical features that are essential for recognition tasks.

Once quantization is complete, the final model is deployed in an offline setting, allowing it to operate efficiently in real-time maritime environments without relying on cloud-based processing. This ensures that ASVs can execute autonomous marine monitoring, environmental analysis, and navigation tasks with minimal computational overhead, making AWQ an essential strategy for deploying deep learning models in practical, edge-computing scenarios.

\section{Experiments}

\subsection{Settings}
\subsubsection{Experimental Environment}
The experiments were conducted in a high-performance computing environment equipped with an AMD Ryzen 9 9950X 16-Core processor (4.30 GHz) and an NVIDIA GeForce RTX 4090 GPU, running Ubuntu 22.04. The deep learning framework used was PyTorch 2.5.1 with Python 3.12 and CUDA 12.1, providing robust computational support for model training and inference.

\subsubsection{Utilized MLLM Model}
For multimodal feature processing, we employed the Llama-3.2-11B-Vision model~\cite{dubey2024llama} to generate textual descriptions and classification vectors. This model integrates both visual and textual feature generation capabilities, providing high-quality prior knowledge for multimodal tasks. The classification vector represents a probabilistic category distribution for each sample, enriching the multimodal fusion input.

\subsubsection{Dataset Partitioning}
We utilized a custom dataset consisting of 500 samples, with 100 samples per category. The categories include marine debris, stranded animals, ship fires, vessel capsizing, and red tides. The dataset was split into training, validation, and test sets in an 8:1:1 ratio, ensuring a balanced category distribution. This partitioning strategy enhances training diversity, improves generalization, and ensures sufficient samples for model validation and testing.

\subsubsection{Comparative Model Setup}
The comparative models were categorized into two groups: purely visual models and multimodal models. The purely visual models included ConvNeXt~\cite{liu_convnet_2022}, ResNet18~\cite{he_deep_2016}, EfficientNet~\cite{tan_efficientnet_2019}, ViT~\cite{dosovitskiy_image_2020}, and Swin Transformer~\cite{liu2021swin}, which represent state-of-the-art deep learning architectures in image classification. These models relied solely on image features for scene classification without incorporating any additional modalities. The multimodal models included CLIP~\cite{radford2021learning} and BLIP~\cite{li2022blip}, two leading approaches in vision-language joint learning, capable of leveraging both visual and textual semantics for classification. 

\subsubsection{Model Training Details}
During training, we adopted the Stochastic Gradient Descent (SGD) optimizer with an initial learning rate of \(1 \times 10^{-3}\) and a momentum factor of 0.9. The model was trained for 30 epochs with a batch size of 8. To mitigate overfitting, dropout regularization with a dropout probability of 0.1 was applied. Prior to fusion, the feature dimensions were adjusted to \(d = 100\). In the Final Decision Layer, the first fully connected layer reduced the feature dimensions from 400 to 50. The second layer projected the 50-dimensional representation into a 5-dimensional output space, corresponding to the 5 classes. A fixed learning rate was maintained, and validation accuracy was monitored to select the best-performing model. After training, we applied 4-bit AWQ. Randomly select 128 images from the 400 training samples as the calibration dataset, ensuring no repetition and unbiased selection. The scaling factors for each channel were determined using the 99th percentile of activation magnitudes from a calibration dataset, with \( \alpha = 0.99 \). This adaptive quantization prioritized high-activation channels for better precision. The model weights were then quantized according to these scaling factors, achieving significant compression while maintaining high performance.

\subsection{Results}

\begin{table}[H]
\centering
\caption{Comparison of Models on Maritime Scene Recognition Task}
\label{tab:comparison}
\renewcommand{\arraystretch}{1.3}
\setlength{\tabcolsep}{3.5pt}
\begin{adjustbox}{width=\linewidth}
\begin{tabular}{l *{7}{c}}
\toprule
\rowcolor[RGB]{240,248,255}
&& \multicolumn{3}{c}{\textbf{Macro Average}} & \multicolumn{3}{c}{\textbf{Weighted Average}} \\
\cline{3-5} \cline{6-8}
\rowcolor[RGB]{240,248,255}
\multirow{-2}{*}{\cellcolor[RGB]{240,248,255}\textbf{Models}} & \multirow{-2}{*}{\cellcolor[RGB]{240,248,255}\textbf{Accuracy}} & \textbf{Precision} & \textbf{Recall} & \textbf{F1} & \textbf{Precision} & \textbf{Recall} & \textbf{F1} \\
\midrule
\rowcolor[RGB]{245,245,255}
\multicolumn{8}{c}{\textbf{Pure Vision Models}} \\
\midrule
ConvNeXt~\cite{liu_convnet_2022} & 92.5 & 92.5 & 92.6 & 92.4 & 92.9 & 92.5 & 92.5 \\
\rowcolor[RGB]{248,248,248}
ResNet18~\cite{he_deep_2016} & 94.5 & 94.5 & 94.8 & 94.5 & 94.7 & 94.5 & 94.5 \\
EfficientNet~\cite{tan_efficientnet_2019} & 91.0 & 91.2 & 91.3 & 91.0 & 91.5 & 91.0 & 91.0 \\
\rowcolor[RGB]{248,248,248}
ViT~\cite{dosovitskiy_image_2020} & 94.0 & 94.2 & 94.2 & 94.1 & 94.2 & 94.0 & 94.0 \\
Swin Transformer~\cite{liu2021swin} & 93.5 & 93.7 & 93.8 & 93.5 & 93.9 & 93.5 & 93.5 \\
\midrule
\rowcolor[RGB]{245,245,255}
\multicolumn{8}{c}{\textbf{Multimodal Models}} \\
\midrule
CLIP-zeroshot~\cite{radford2021learning} & 86.4 & 89.8 & 86.4 & 85.1 & 89.8 & 86.4 & 85.1 \\
\rowcolor[RGB]{248,248,248}
BLIP-zeroshot~\cite{li2022blip} & 92.4 & 93.5 & 92.4 & 92.5 & 93.5 & 92.4 & 92.5 \\
CLIP~\cite{radford2021learning} & 93.5 & 94.0 & 94.5 & 94.3 & 94.5 & 94.3 & 94.0 \\
\rowcolor[RGB]{248,248,248}
BLIP~\cite{li2022blip} & 94.5 & 95.2 & 94.5 & 95.2 & 95.0 & 95.3 & 94.3 \\
\rowcolor[RGB]{230,240,255}
\textbf{Proposed Model (Ours)} & \textbf{98.0} & \textbf{98.0} & \textbf{98.2} & \textbf{98.0} & \textbf{98.1} & \textbf{98.0} & \textbf{98.0} \\
\bottomrule
\end{tabular}
\end{adjustbox}
\end{table}

As shown in Table~\ref{tab:comparison}, the comparison of the Pure Vision Models and the Proposed Model demonstrates a clear improvement in performance across all metrics. The Pure Vision Models, such as ConvNeXt~\cite{liu_convnet_2022}, ResNet18~\cite{he_deep_2016}, EfficientNet~\cite{tan_efficientnet_2019}, ViT~\cite{dosovitskiy_image_2020}, and Swin Transformer~\cite{liu2021swin}, exhibit strong accuracy and precision, with values ranging from 91.0$\%$ to 94.5$\%$. However, these models rely solely on image data for feature extraction, which limits their ability to capture the full contextual information available in complex scenarios, such as maritime scene recognition. Our proposed model, which integrates image, text, and vector modalities, achieves a remarkable accuracy.

In comparison with multimodal models like CLIP~\cite{radford2021learning} and BLIP~\cite{li2022blip}, our Proposed Model (Ours) achieves superior performance across all evaluation metrics. While CLIP and BLIP perform well with accuracy rates of 93.5$\%$ and 94.5$\%$, respectively, they do not fully exploit the potential of integrating features from diverse modalities. Our model builds upon the strengths of multimodal learning by using a more advanced fusion strategy, which includes not only visual and textual data but also carefully integrated classification vectors. This approach results in a 98.0$\%$ accuracy, accompanied by improved macro-average and weighted average precision, recall, and F1 scores.

After applying AWQ, our model's accuracy slightly decreased by only 0.5$\%$, resulting in an accuracy of 97.5$\%$. Despite this small reduction, AWQ enabled a significant reduction in model size and computational overhead, making it suitable for real-time deployment in resource-constrained maritime environments. The enhanced fusion strategy, coupled with dynamic modality prioritization and mutual information maximization, continues to provide our model with a more comprehensive understanding of the maritime scene, ensuring that each modality contributes optimally to the final prediction.

\subsection{Analysis}
\subsubsection{Quantization Impact}

In this section, we analyze the effects of AWQ on the model's performance and resource utilization. AWQ significantly reduces the model size and computational overhead, making it more suitable for deployment in resource-constrained environments. However, it also causes a slight decrease in accuracy, which is typical for most quantization techniques. Below, we summarize the key changes in both model performance and resource usage before and after AWQ quantization.

\begin{table}[H]
\centering
\caption{Impact of AWQ Quantization on Model Performance and Resource Utilization}
\label{tab:quantization_impact}
\renewcommand{\arraystretch}{1.3}
\setlength{\tabcolsep}{3.5pt}
\begin{adjustbox}{width=\linewidth}
\begin{tabular}{l c c}
\toprule
\rowcolor[RGB]{240,248,255}
\textbf{Metric} & \textbf{Proposed Model (Full-Precision)} & \textbf{Proposed Model (AWQ-4bit)} \\
\midrule
\rowcolor[RGB]{245,245,255}
\multicolumn{3}{c}{\textbf{Performance Metrics}} \\
\midrule
\textbf{Accuracy} & 98.0 & 97.5 \\
\rowcolor[RGB]{248,248,248}
\textbf{Precision (Macro Average)} & 98.0 & 97.3 \\
\textbf{Recall (Macro Average)} & 98.2 & 97.5 \\
\rowcolor[RGB]{248,248,248}
\textbf{F1 (Macro Average)} & 98.0 & 97.4 \\
\midrule
\rowcolor[RGB]{245,245,255}
\multicolumn{3}{c}{\textbf{Resource Utilization}} \\
\midrule
\textbf{FLOPs (G)} & 28.5 & 28.5 \\
\rowcolor[RGB]{248,248,248}
\textbf{Params (M)} & 137.5 & 137.5 \\
\textbf{Model Size (MB)} & 550 & \textbf{68.75} \\
\rowcolor[RGB]{248,248,248}
\textbf{Peak Mem. (GB)} & 5.5 & \textbf{3.2} \\
\textbf{Batch Time (ms)} & 15 & \textbf{5.2} \\
\rowcolor[RGB]{248,248,248}
\textbf{Throughput (img/s)} & 533.3 & \textbf{1538.5} \\
\bottomrule
\end{tabular}
\end{adjustbox}
\end{table}

Table~\ref{tab:quantization_impact} presents a comparison of the model's performance and resource utilization metrics before and after applying AWQ. Applying AWQ with 4-bit quantization results in a slight accuracy drop of 0.5$\%$, from 98.0$\%$ to 97.5$\%$ However, the reduction in model size from 550 MB to 68.75 MB and the significant improvement in throughput, from 533.3 to 1538.5, demonstrate the efficiency of AWQ. Peak memory usage also decreases from 5.5 GB to 3.2 GB, and batch time is reduced from 15 ms to 5.2 ms, making the quantized model more suitable for real-time deployment in resource-constrained environments, such as ASVs. The number of floating-point operations (FLOPs) and model parameters remain unchanged, as AWQ primarily focuses on optimizing weight precision without altering the network architecture.

In conclusion, while AWQ results in a slight drop in accuracy, the trade-off is highly favorable, as it significantly reduces the model's memory and computational requirements, making it a practical solution for deployment in real-world maritime applications.

\subsubsection{Ablation Study}
\begin{table}[H]
\caption{Ablation Study for Multimodal Architectures on Maritime Scene Recognition}
\label{tab:model_comparison}
\renewcommand{\arraystretch}{1.3}
\centering
\setlength{\tabcolsep}{3.5pt}
\begin{adjustbox}{width=\linewidth}
\begin{tabular}{l *{7}{c}}
\toprule
\rowcolor[RGB]{245,245,255}
& & \multicolumn{3}{c}{\textbf{Macro Average}} & \multicolumn{3}{c}{\textbf{Weighted Average}} \\
\cline{3-5} \cline{6-8}
\rowcolor[RGB]{245,245,255}
\multirow{-2}{*}{\cellcolor[RGB]{245,245,255}\textbf{Model Architecture}} & \multirow{-2}{*}{\cellcolor[RGB]{245,245,255}\textbf{Accuracy}} & \textbf{Precision} & \textbf{Recall} & \textbf{F1} & \textbf{Precision} & \textbf{Recall} & \textbf{F1} \\
\midrule
\rowcolor[RGB]{230,240,255}
\textbf{Swin-T + BERT} (Ours) & \textbf{98.0} & \textbf{98.0} & \textbf{98.2} & \textbf{98.0} & \textbf{98.1} & \textbf{98.0} & \textbf{98.0} \\
\rowcolor[RGB]{248,248,248}
\textbf{ViT + BERT} & 97.5$^{\textcolor{red}{\downarrow}0.5}$ & 97.6$^{\textcolor{red}{\downarrow}0.4}$ & 97.6$^{\textcolor{red}{\downarrow}0.6}$ & 97.6$^{\textcolor{red}{\downarrow}0.4}$ & 97.6$^{\textcolor{red}{\downarrow}0.5}$ & 97.5$^{\textcolor{red}{\downarrow}0.5}$ & 97.5$^{\textcolor{red}{\downarrow}0.5}$ \\
\textbf{ViT + BERT*} & 97.0$^{\textcolor{red}{\downarrow}0.5}$ & 96.9$^{\textcolor{red}{\downarrow}0.5}$ & 97.2$^{\textcolor{red}{\downarrow}0.4}$ & 97.0$^{\textcolor{red}{\downarrow}0.6}$ & 97.2$^{\textcolor{red}{\downarrow}0.4}$ & 97.0$^{\textcolor{red}{\downarrow}0.5}$ & 97.0$^{\textcolor{red}{\downarrow}0.5}$ \\
\rowcolor[RGB]{248,248,248} 
\textbf{ViT + CNN*} & 88.3$^{\textcolor{red}{\downarrow}8.7}$ & 90.2$^{\textcolor{red}{\downarrow}6.7}$ & 88.9$^{\textcolor{red}{\downarrow}8.3}$ & 87.5$^{\textcolor{red}{\downarrow}9.5}$ & 91.0$^{\textcolor{red}{\downarrow}6.2}$ & 88.2$^{\textcolor{red}{\downarrow}8.8}$ & 88.1$^{\textcolor{red}{\downarrow}8.9}$ \\
\bottomrule
\end{tabular}
\end{adjustbox}
\vspace{1mm}
\raggedright
\small Note: * indicates model without multimodal fusion. All models include MLP component.
\end{table}

To validate the effectiveness of the proposed model, we conducted an ablation study to assess the impact of different configurations on maritime scene recognition. As shown in Table~\ref{tab:model_comparison}, the results clearly demonstrate that our model, which combines Swin Transformer for image processing, BERT for text processing, MLP for classification vector handling, and a multimodal fusion strategy, outperforms all other configurations across all evaluation metrics.

Our model achieved an accuracy of 98.0$\%$, with a macro-average F1 score of 98.0$\%$, and macro-average precision and recall of 98.0$\%$ and 98.2$\%$, respectively. The weighted average precision, recall, and F1 score were 98.1$\%$, 98.0$\%$, and 98.0 $\%$, respectively, indicating significant performance gains through multimodal fusion. The incorporation of information from image, text, and classification vectors greatly enhanced recognition accuracy, particularly in complex maritime scenarios.

Among these, the ViT + CNN configuration performed particularly poorly, with an accuracy of only 88.3$\%$, reflecting a major performance drop due to the lack of effective multimodal fusion. This reinforces the significance of multimodal strategies for improving model robustness and accuracy in real-world applications.

\begin{table}[H]
\centering
\caption{Ablation Study for Fusion Strategies on Maritime Scene Recognition}
\label{tab:ablation}
\renewcommand{\arraystretch}{1.3}
\setlength{\tabcolsep}{3.5pt}
\begin{adjustbox}{width=\linewidth}
\begin{tabular}{l *{7}{c}}
\toprule
\rowcolor[RGB]{245,245,255}
& & \multicolumn{3}{c}{\textbf{Macro Average}} & \multicolumn{3}{c}{\textbf{Weighted Average}} \\
\cline{3-5} \cline{6-8}
\rowcolor[RGB]{245,245,255}
\multirow{-2}{*}{\cellcolor[RGB]{245,245,255}\textbf{Fusion Strategy}} & \multirow{-2}{*}{\cellcolor[RGB]{245,245,255}\textbf{Accuracy}} & \textbf{Precision} & \textbf{Recall} & \textbf{F1} & \textbf{Precision} & \textbf{Recall} & \textbf{F1} \\
\midrule
\rowcolor[RGB]{230,240,255}
\textbf{Complete Fusion Strategy (Ours)} & \textbf{98.0} & \textbf{98.0} & \textbf{98.2} & \textbf{98.0} & \textbf{98.1} & \textbf{98.0} & \textbf{98.0} \\
\rowcolor[RGB]{248,248,248}
Stacking (Image + Text + Vector) & 97.2 & 97.3 & 97.4 & 97.2 & 97.4 & 97.1 & 97.2 \\
Attention-based Fusion & 97.8 & 97.9 & 98.0 & 97.8 & 98.0 & 97.8 & 97.9 \\
\rowcolor[RGB]{248,248,248}
Weighted Integration (No Attention) & 97.5 & 97.6 & 97.7 & 97.5 & 97.7 & 97.5 & 97.6 \\
Enhanced Modal Alignment & 97.9 & 98.0 & 98.1 & 97.9 & 98.0 & 97.9 & 98.0 \\
\bottomrule
\end{tabular}
\end{adjustbox}
\vspace{1mm}
\raggedright
\small Note: This ablation study performed on the Swin + BERT + MLP architecture.
\end{table}

Table~\ref{tab:ablation} presents an ablation study on various fusion strategies for maritime scene recognition. The results show that the Complete Fusion Strategy achieves the highest accuracy of 98.0$\%$, surpassing all other strategies in both macro and weighted averages across precision, recall, and F1 score. This indicates that the full integration of image, text, and vector modalities with attention mechanisms yields the best overall performance.

Other fusion strategies, such as Stacking (Image + Text + Vector), Attention-based Fusion, and Enhanced Modal Alignment, demonstrate lower accuracy. Specifically, Stacking produces the lowest performance, while Attention-based Fusion and Enhanced Modal Alignment perform slightly better but still fall short of the complete fusion approach. The Weighted Integration (No Attention) strategy shows relatively balanced performance but also lags behind the complete fusion strategy.

\begin{table}[H]
\centering
\caption{Ablation Study for Between Pure and Enhanced MLLM on Maritime Scene Recognition}
\label{tab:mllm_comparison}
\renewcommand{\arraystretch}{1.3}
\setlength{\tabcolsep}{3.5pt}
\begin{adjustbox}{width=\linewidth}
\begin{tabular}{l *{7}{c}}
\toprule
\rowcolor[RGB]{245,245,255}
& & \multicolumn{3}{c}{\textbf{Macro Average}} & \multicolumn{3}{c}{\textbf{Weighted Average}} \\
\cline{3-5} \cline{6-8}
\rowcolor[RGB]{245,245,255}
\multirow{-2}{*}{\cellcolor[RGB]{245,245,255}\textbf{MLLM Architecture}} & \multirow{-2}{*}{\cellcolor[RGB]{245,245,255}\textbf{Accuracy}} & \textbf{Precision} & \textbf{Recall} & \textbf{F1} & \textbf{Precision} & \textbf{Recall} & \textbf{F1} \\
\midrule
\rowcolor[RGB]{248,248,248}
\textbf{Pure MLLM} & 88.0 & 90.3 & 88.4 & 87.8 & 90.3 & 88.4 & 87.8 \\
\rowcolor[RGB]{230,240,255}
\textbf{Enhanced MLLM} (Ours) & \textbf{98.0} & \textbf{98.0} & \textbf{98.2} & \textbf{98.0} & \textbf{98.1} & \textbf{98.0} & \textbf{98.0} \\
\bottomrule
\end{tabular}
\end{adjustbox}
\end{table}

Finally, we evaluated the classification vector generated by the MLLM model alone. As shown in Table~\ref{tab:mllm_comparison}, its performance was relatively low, with an accuracy of 88.0$\%$, and other metrics did not reach the level of the multimodal models. This suggests that while classification vectors provide relatively accurate class probability information, they still require integration with other modalities to address the complexity of maritime scene recognition effectively.

In summary, the results confirm that our framework provides a substantial performance advantage over other configurations, highlighting the importance of integrating diverse modalities for optimal recognition performance.

\subsubsection{Computing Efficiency and Resource Utilization Analysis} 
\begin{table}[H]
\centering
\caption{Computational Efficiency and Resource Utilization Analysis}
\label{tab:efficiency_analysis}
\renewcommand{\arraystretch}{1.3}
\setlength{\tabcolsep}{3.5pt}
\begin{adjustbox}{width=\linewidth}
\begin{tabular}{l *{6}{c}}
\toprule
\rowcolor[RGB]{245,245,255}
& \textbf{FLOPs} & \textbf{Params} & \textbf{Peak Mem.} & \textbf{Batch} & \textbf{Throughput} & \textbf{Train} \\
\rowcolor[RGB]{245,245,255}
\multirow{-2}{*}{\cellcolor[RGB]{245,245,255}\textbf{Model}} & \textbf{(G)} & \textbf{(M)} & \textbf{(GB)} & \textbf{Time (ms)} & \textbf{(img/s)} & \textbf{Time (h)} \\
\midrule
ResNet18~\cite{he_deep_2016} & 1.8 & 11.7 & 46.8 & 1.1 & 3.0 & 2666.7 \\
\rowcolor[RGB]{248,248,248}
ViT~\cite{dosovitskiy_image_2020} & 17.6 & 86.4 & 345.6 & 2.8 & 6.5 & 1230.8 \\
Swin Transformer~\cite{liu2021swin} & 4.5 & 28 & 112 & 1.6 & 4.0 & 2000.0 \\
\rowcolor[RGB]{248,248,248}
BLIP~\cite{li2022blip} & 24.3 & 92.3 & 369.2 & 3.5 & 8.0 & 1000.0 \\
\midrule
\rowcolor[RGB]{230,240,255}
Proposed Model (AWQ-4bit) & \textbf{28.5} & \textbf{137.5} & \textbf{68.75} & \textbf{3.2} & \textbf{5.2} & \textbf{1538.5} \\
\bottomrule
\end{tabular}
\end{adjustbox}
\vspace{1mm}
\raggedright
\small \textit{Note}: All metrics measured on NVIDIA RTX 4090.
\end{table}

As shown in Table~\ref{tab:efficiency_analysis}, we compare the computational efficiency and resource utilization of the Proposed Model (AWQ-4bit) with other baseline models. It is important to note that the metrics in this analysis are measured during inference and not training.

Our Proposed Model (AWQ-4bit), after applying 4-bit Activation-Aware Quantization, significantly reduces the model size to 68.75MB, while maintaining an impressive inference throughput of 1538.5 images per second. Although the Proposed Model (AWQ-4bit) has a slightly lower throughput compared to the ResNet18 (2666.7 images/s), it still achieves superior accuracy for maritime scene recognition, with a minimal drop in performance (only 0.5$\%$) after quantization.

In terms of memory usage, Proposed Model (AWQ-4bit) shows a peak memory utilization of 3.2GB, which is a significant improvement compared to other large models like BLIP (3.5GB). These results demonstrate that our quantized model is highly optimized for deployment in resource-constrained environments such as ASVs, where computational resources are limited.

While there is still a performance gap between our model and the baseline models in terms of throughput, the Proposed Model (AWQ-4bit) strikes a strong balance between accuracy, model size, and computational efficiency. It achieves one of the highest recognition accuracies in this category, demonstrating its suitability for real-time maritime scene recognition tasks on ASVs, where both efficiency and accuracy are critical. The trade-off in throughput is minimal, given the model's competitive performance in all other aspects.

Overall, AWQ quantization has proven to be an effective lightweight solution for enhancing the deployment feasibility of deep learning models in edge computing systems, particularly for maritime applications where both resource utilization and recognition performance are paramount.

\subsubsection{Validation on Challenging Samples}

To further evaluate the robustness and practical applicability of our proposed model after applying AWQ, we conducted validation using a set of challenging images from the test dataset, each representing a different marine scene category. These images were selected for their significant complexities, such as low resolution, target occlusion, complex backgrounds, and interference from environmental elements, to simulate real-world operational difficulties.

\begin{figure}[H]
\centering
\includegraphics[width=\linewidth]{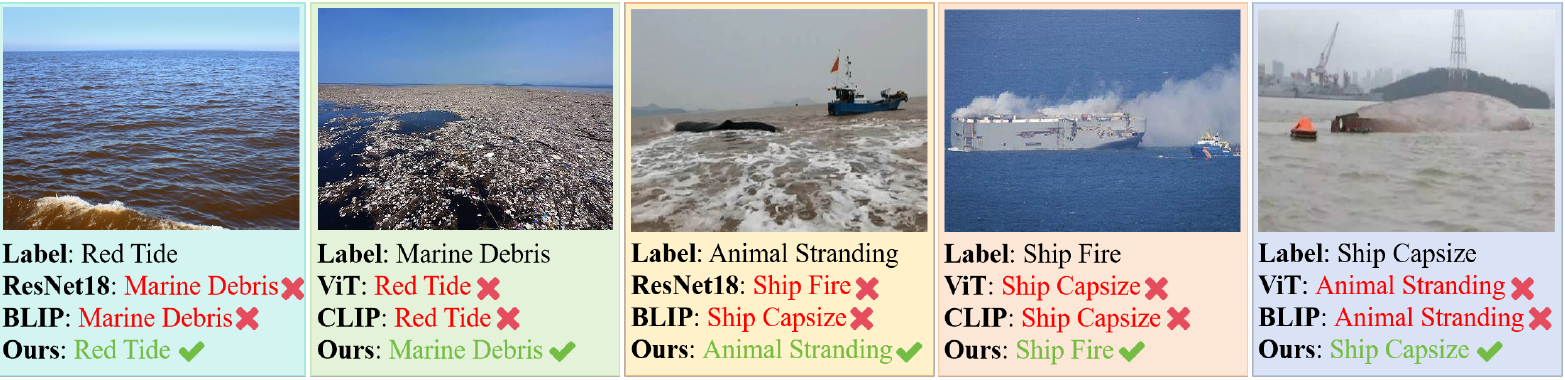}
\caption{Challenging maritime scene samples showing correct classifications by our model. The true labels are displayed, with our model's accurate predictions in green and the incorrect predictions from other models in red.}\label{validation}
\end{figure}

As shown in the ~\figref{validation}, our proposed model demonstrated strong performance on several challenging maritime scene samples, accurately identifying the correct labels even in the presence of difficult visual cues. For example, in the first image depicting Red Tide, the water surface color is relatively dark, making it challenging to distinguish it from Marine Debris. While both ResNet18 and BLIP misclassify it as Marine Debris, our model successfully classifies it as Red Tide. This can be attributed to our model's effective multimodal fusion strategy, which integrates image, text, and classification vectors to enhance the recognition of subtle differences between similar scenes.

In the third image showing a stranded animal, there are visual distractions such as the presence of a boat, which could easily mislead the model into classifying it as Ship Capsize, as seen in the incorrect predictions from ViT and BLIP. However, our model, which leverages the integration of multiple modalities and the power of MLLM, is able to accurately identify the scene as Animal Stranding. This demonstrates how the multimodal approach combined with MLLM enhances the model's robustness in complex situations and improves its ability to focus on relevant features for accurate classification.


\section{Conclusions}
This study presents a novel multimodal AI framework for marine environment monitoring, seamlessly integrating image analysis, textual annotations, and classification vectors. The proposed model achieves an impressive accuracy of 98.0$\%$, demonstrating the effectiveness of its advanced multimodal fusion strategy. To further optimize deployment efficiency, we apply AWQ, reducing the model size to 68.75MB while maintaining a 97.5$\%$ accuracy—only a minor 0.5$\%$ drop from the full-precision model. This compression significantly enhances computational efficiency, making the system particularly suitable for deployment in resource-constrained environments, such as ASVs. With its robust performance, this framework excels in marine scene classification, providing high-performance solutions for real-time applications, including environmental monitoring and navigation safety.

Our approach not only offers a significant improvement in classification accuracy but also demonstrates the feasibility of deploying complex deep learning models on edge devices with limited computational resources. The model’s ability to efficiently process multimodal data in real-time is crucial for applications in dynamic marine environments, where computational efficiency and model robustness are essential.

Future work will focus on expanding the dataset to incorporate a broader range of marine scenarios, ensuring that the model can handle diverse real-world conditions. Additionally, conducting field experiments to validate the model’s performance in dynamic and unpredictable marine environments will be key to assessing its practical viability. To further improve the model’s scalability, we plan to explore semi-supervised learning techniques and adaptive processing strategies, which will enhance its generalization capabilities and reduce dependency on labeled data. These efforts will help fine-tune the model’s performance and make it more adaptable to various maritime tasks. Ultimately, our goal is to optimize the model for real-world deployment, contributing to marine conservation, operational safety, and other critical maritime applications.

\appendix

\section{Prompts for Textual Description and Classification Vector Generation}

This appendix contains the prompts used for generating textual descriptions and classification vectors for the maritime scene recognition task. These prompts were designed to provide semantic context and probabilistic classifications for the images used in this study.

\subsection{Prompt for Textual Description Generation}
\begin{promptbox}[Prompt for Textual Description Generation]{c1}
You are given an image. Create a clear, concise description (approximately 150 words).

- Image: \{\}

Your description must:

1. Identify and describe the main elements in the image

   - Primary subjects/objects
   
   - Setting/background
   
   - Colors and visual characteristics
   
   - Actions or interactions occurring

2. Follow this structure:

   a) Brief overview (1 sentence)
   
   b) Key details (1-2 sentences)
   
Be objective, precise, and focus only on what is clearly visible.
Prioritize clarity over elaboration.

Provide only the description without additional commentary.
\end{promptbox}

\subsection{Prompt for Classification Vector Generation}

\begin{promptbox}[Prompt for Classification Vector Generation]{c1}
Analyze the provided image and classify it into one of the following categories:

- Image: \{\}

Categories:

1. 'Red Tide'

2. 'Marine Debris' 

3. 'Animal Stranding'

4. 'Ship Fire'

5. 'Ship Capsize'

Instructions:

1. Examine the image carefully for visual cues relevant to each category

2. Assess the probability that the image belongs to each category

3. Assign probability values that sum to exactly 1.0

Output format:

Provide your classification as a probability distribution vector:

$[p_1, p_2, p_3, p_4, p_5]$

Where:

- $p_1$ = probability of 'Red Tide'

- $p_2$ = probability of 'Marine Debris'

- $p_3$ = probability of 'Animal Stranding'

- $p_4$ = probability of 'Ship Fire'

- $p_5$ = probability of 'Ship Capsize'

Example output: $[0.1, 0.5, 0.2, 0.1, 0.1]$

Return only the probability vector without additional explanation.

\end{promptbox}
\bibliographystyle{unsrt} 
\bibliography{references}  






\end{document}